\documentclass[11pt]{article}

\usepackage{fullpage}
\usepackage{amsfonts,epsfig,graphicx}
\usepackage{amsmath,amssymb,amsthm}
\usepackage{graphics}
\usepackage{macros}
\usepackage[numbers]{natbib}
\usepackage{color}
\usepackage[ruled]{algorithm2e}
\usepackage{epstopdf}
\usepackage{algorithmic}
\usepackage{enumerate}
\usepackage[bookmarks=true,colorlinks,citecolor=blue,urlcolor=blue]{hyperref}

\newtheorem{theorem}{Theorem}
\newtheorem{lemma}{Lemma}

\newtheorem{definition}{Definition}
\newtheorem{proposition}{Proposition}

\title{$\ell_1$-regularized Neural Networks are Improperly Learnable\\ in Polynomial Time}
\author{Yuchen Zhang\qquad Jason D. Lee\qquad Michael I. Jordan\\
Department of Electrical Engineering and Computer Science\\
University of California, Berkeley, CA 94720\\
\tt \{yuczhang,jasondlee,jordan\}@eecs.berkeley.edu}
\date{}

\begin{document}

\maketitle

\begin{abstract}
\noindent 
We study the improper learning of multi-layer neural networks. Suppose that the neural network to be learned has $k$ hidden layers and that the $\ell_1$-norm of the incoming weights of any neuron is bounded by $L$. We present a kernel-based method, such that with probability at least $1 - \delta$, it learns a predictor whose generalization error is at most $\epsilon$ worse than that of the neural network. The sample complexity and the time complexity of the presented method are polynomial in the input dimension and in $(1/\epsilon,\log(1/\delta),F(k,L))$, where $F(k,L)$ is a function depending on $(k,L)$ and on the activation function, independent of the number of neurons. The algorithm applies to both sigmoid-like activation functions and ReLU-like activation functions. It implies that any sufficiently sparse neural network is learnable in polynomial time. 
\end{abstract}


\section{Introduction}

Neural networks have been successfully applied in many areas of artificial intelligence, such as image classification, face recognition, speech recognition and natural language processing. Practical successes have been driven by the rapid growth in the size of data sets and the increasing availability of large-scale parallel and distributed computing platforms. Examples of recent work in this area include~\cite{le2013building, krizhevsky2012imagenet, zeiler2014visualizing, chen2014fast, dahl2013improving, hinton2012deep}.

The theoretical understanding of learning in neural networks has lagged the practical successes. It is known that any smooth function can be approximated by a network with just one hidden layer~\cite{barron1993universal}, but training such a  network is NP-hard~\cite{blum1992training}. In practice, people use optimization algorithms such as stochastic gradient descent (SGD) to train neural networks. Although strong theoretical results are available for SGD in the setting of convex objective functions, there are few such results in the nonconvex setting of neural networks. While it is possible to transform the neural network training problem to a convex optimization problem involving an infinite number of variables~\cite{bengio2005convex}, the infinitude of variables means that there is no longer a guarantee that the learning algorithm will terminate in polynomial time.

Several recent papers have risen to the challenge of establishing polynomial-time learnability results for neural networks.  These papers necessarily (given that the problem is NP-hard) introduce additional assumptions or relaxations. For instance, one may assume that the data is in fact generated by the neural network. Under this assumption,
Arora et al.~\citep{arora2013provable} study the recovery of denoising auto-encoders which are represented by multi-layer neural networks.
They assume that the top-layer values of the network are randomly generated and all network weights are randomly drawn from $\{-1,1\}$. As a consequence, the bottom layer generates a sequence of random observations using which the algorithm can recover the network weights. The algorithm has polynomial-time complexity and is capable of learning random networks that are drawn from a specific distribution. However, in practice people want to learn deterministic networks that encode data-dependent representations.

Sedghi and Anandkumar~\cite{sedghi2014provable} study the supervised learning of neural networks under the assumption that the data distribution has a score function that is known in advance. They show that if the input dimension is large enough and the network is sparse enough, then the first network layer can be learned by a polynomial-time algorithm. Learning the deeper layers remains as an open problem. In addition, their method assumes that the network weights are randomly drawn from a Bernoulli-Gaussian distribution. More recently, Janzamin et al.~\cite{janzamin2015generalization} propose another algorithm based on the score function that removes the restrictions of Sedghi and Anandkumar~\cite{sedghi2014provable}. The assumption in this case is that the network weights satisfy a non-degeneracy condition; moreover, the algorithm is only capable of learning neural networks with one hidden layer. 

Another approach to the problem is via the improper learning framework. The goal in this case is to find a predictor that is not a neural network, but performs as well as the best possible neural network in terms of the generalization error.
Livni et al.~\cite{livni2014computational} consider changing the activation function and over-specifying the network to make it easier to train. They show that polynomial networks (e.g., networks whose activation function is quadratic) with sufficient width and depth are as expressive as the sigmoid-activated neural networks. Although a deep polynomial network is still hard to train, they propose training in a superclass---the class of all polynomial functions with bounded degree. As a consequence, there is an improper learning algorithm which achieves a generalization error at most $\epsilon$ worse than that of the best neural network. The time complexity is polynomial in the input dimension $d$ and quasi-polynomial in $1/\epsilon$. Since the dependence on $d$ has a large power, the algorithm is not practical unless $d$ is quite small. Livni et al.~\cite{livni2014computational} further show, however, that there is a practical algorithm to directly train the polynomial network if it has one or two hidden layers. 

A recent line of work has focused on understanding the energy landscape of a neural network. After several simplifying assumptions, a neural network can be shown to be a Gaussian field whose critical points can be analyzed using the Kac-Rice formula and  properties  of the Gaussian Orthogonal Ensemble \cite{auffinger2013random,dauphin2014identifying,choromanska2014loss}. The conclusion of these papers is that all critical points with nonnegative eigenvalues tend to have objective value near the global minimum. Thus in such networks if we could find such a point, it would have small objective value and thus small training error. This combined with generalization error bounds would imply finding a neural network with low excess risk. However, there is no provably efficient algorithm for finding a critical point with nonnegative eigenvalues.

\subsection{Our contribution}

In this paper, we propose a practical algorithm called the \emph{recursive kernel method} for learning multi-layer neural networks, under the framework of improper learning. 
Our method is inspired by the work of Shalev-Shwartz et al.~\cite{shalev2011learning}, which shows that for binary classification with the sigmoidal loss,
there is a kernel-based method that achieves the same generalization error as the best linear classifier. We extend this method to deeper networks. In particular, we assume that the neural network to be learned takes $d$-dimensional input. It has $k$ hidden layers and the $\ell_1$-norm of the incoming weights of any neuron is bounded by $L$. Under these assumptions, the algorithm learns a kernel-based predictor whose generalization error is at most $\epsilon$ worse than that of the best neural network. The sample and the time complexity of the algorithm are polynomial in $(d,1/\epsilon,\log(1/\delta),F(k,L))$, where $F(k,L)$ is a function depending on $(k,L)$ and on the activation function, independent of the input dimension or the number of neurons. The theoretical result holds for any data distribution.

As concrete examples, we demonstrate that if the activation function is a quadratic function, then $F(k,L)$ is a polynomial function of $L$. Thus, the algorithm recovers the theoretical guarantee of Livni et al.~\cite{livni2014computational}. We also demonstrate two activation functions, one that approximates the sigmoid function and the other that approximates the ReLU function, under which $F(k,L)$ is finite. Thus, the algorithm also learns neural networks activated by sigmoid-like or ReLU-like functions.
For these latter examples, the dependence on $L$ is no longer polynomial. This non-polynomial dependence is in fact inevitable: Under a hardness assumption in cryptographics and assuming sigmoid-like or ReLU-like activation, we prove that no algorithm running in  $\poly(L)$ time can improperly learn the neural network.

The paper is organized as follow. In Section~\ref{sec:setup}, we formalize the problem and clarify the assumptions that we make for the theoretical analysis. In Section~\ref{sec:alg-and-theory}, the algorithm and the associated theoretical results are presented. We discuss concrete examples to demonstrate the application of the theory.
In Section~\ref{sec:hardness}, we present hardness results for the improper learning of neural networks.
In Section~\ref{sec:experiment}, we report experiments on the MNIST dataset and its variations, demonstrating that in addition to its role in our theoretical analysis the proposed algorithm is comparable in practice with baseline neural network learning methods.

\section{Problem Setup}
\label{sec:setup}

We consider a fully-connected neural network $\nn$ that maps a vector $x\in\R^d$ to a real number $\nn(x)$ via $k$ hidden layers. Let $d^{(p)}$ represent the number of neurons in the $p$-th layer. Let $y^{(p)}_i$ represent the output of the $i$-th neuron in the $p$-th layer. We define the zero-th layer to be the input vector so that $d^{(0)} =d$ and $y^{(0)} = x$. The transformation performed by the neural network is defined as follows:
\[
	y^{(p)}_i \defeq \sigma\Big( \sum_{j=1}^{d^{(p-1)}} w^{(p-1)}_{i,j} y^{(p-1)}_j \Big)
	\quad \mbox{and} \quad \nn(x) \defeq \sum_{j=1}^{d^{(k)}} w^{(k)}_{1,j} y^{(k)}_j,
\]
where $w^{(p-1)}_{i,j}$ is the weight of the edge that connects the neuron $j$ on the $(p-1)$-th layer to the neuron $i$ on the $p$-th layer. The activation function $\sigma: \R\to \R$ is a one-dimensional nonlinear function. We will 
discuss the choice of function $\sigma$ later in this section. 

We assume that the input vector has bounded $\ell_2$-norm and the edge weights have bounded $\ell_1$ or $\ell_2$ norms. The assumptions are formalized as follows.

\begin{assumption}\label{assu:boundedness}
The input vector $x$ satisfies $\ltwos{x}\leq 1$. The neuron edge weights satisfy
\begin{align*}
\sum_{j=1}^{d} (w_{i,j}^{(0)})^2 &\leq L^2 \quad \mbox{for all } i \in \{1,\dots,d\}.\\
\sum_{j=1}^{d^{(p)}} |w_{i,j}^{(p)}| &\leq L \quad \mbox{for all } (p,i) \in \{1,\dots,k\} \times \{1,\dots, d^{(p+1)}\}.
\end{align*}
Let $\nn_{k,L,\sigma}$ be the set of $k$-layer neural networks with activation function $\sigma$ that satisfy the edge weight constraints.
\end{assumption}

Assumption~\ref{assu:boundedness} implies that for all neurons on the first hidden layer, the $\ell_2$-norm of their incoming weights is bounded by $L$. For other neurons, the $\ell_1$-norm of their incoming weights is bounded by~$L$. The $\ell_1$-regularization imposes sparsity on the neural network. It is observed in practice that sparse neural networks are capable of learning meaningful representations. For example, the convolution neural network has sparse edges. It has been argued that sparse connectivity is a natural constraint which can lead to improved performance in practice~\cite{thom2013sparse}.

In a prediction task, there is a convex function $\ell: \R\times\R\to \R$ that measures the loss of the prediction. For a feature-label pair $(x,y)\in \mathcal{X}\times R$, its prediction loss is measured by $\ell(\nn(x),y)$. We assume that $(x,y)$ is sampled from an underlying distribution $\mathcal{D}$. The prediction risk of the neural network is defined by $\E[\ell(\nn(x),y)]$. Our goal is to learn a predictor $f:\mathcal{X}\to \R$, which is not necessarily a neural network, such that
\begin{align}\label{eqn:epsilon-optimality-bound}
	\E[\ell(f(x),y)] \leq\arg\min_{\nn \in \nn_{k,L,\sigma}} \E[\ell(\nn(x),y)] + \epsilon. 
\end{align}
In other words, we want to learn a predictor whose generalization loss is at most $\epsilon$ worse than that of the best neural network in $\nn_{k,L,\sigma}$.

In practice, both the sigmoid function $\sigma(x) = (1+e^{-\beta x})^{-1}$ and the ReLU
function $\sigma(x) = \max(0,x)$ are widely used as activation functions for neural networks. We define two classes of activation functions that includes the sigmoid and ReLU respectively.

\begin{definition}[sigmoid-like activation]
A function $\sigma$ is called sigmoid-like if it is non-decreasing on $(-\infty,+\infty)$ and
\[
	\lim_{x\to-\infty} x^c\sigma(x) = 0 \quad \mbox{and} \quad 
	\lim_{x\to \infty} x^c(1 - \sigma(x)) = 0
\]
for some positive constant $c$.
\end{definition}

\begin{definition}[ReLU-like activation]
A function $\sigma$ is called ReLU-like if $\sigma(x) - \sigma(x-1)$ a 
sigmoid-like function.
\end{definition}

Intuitively, a sigmoid-like function is a non-decreasing function on $[0,1]$. When $x\to -\infty$ or $x\to \infty$, the function value approaches $0$ or $1$ at a polynomial rate (or faster) in $x$. A ReLU-like function is a convex function on $[0,\infty)$. When $x\to\infty$, it approaches a linear function with unit slope.

\section{Algorithm and Theoretical Result}
\label{sec:alg-and-theory}

In this section, we present a kernel method which learns a predictor performing as well as the neural network.
We begin by recursively defining a sequence of kernels. Let $K:\R^\N \times \R^\N \to \R$ be a function defined by
\[
	K(x,y) \defeq \frac{1}{2 - \langle x, y \rangle},
\]
where both $\ltwos{x}$ and $\ltwos{y}$ are assumed to be bounded by one.
The function $K$ is a kernel function because we can find a mapping $\psi: \R^\N \to \R^\N$ such that $K(x,y) = \langle \psi(x),\psi(y) \rangle$. The function $\psi$ maps an infinite-dimensional vector to an infinite-dimensional vector. We use $x_i$ to represent the $i$-th coordinate of an infinite-dimensional vector $x$. The $(k_1,\dots,k_j)$-th coordinate of $\psi(x)$, where $j\in \N$ and $k_1,\dots,k_j\in \N$, is defined as $2^{-\frac{j+1}{2}}x_{k_1}\dots x_{k_j}$. By this definition, we have
\begin{align}\label{eqn:psi-product}
	\langle \psi(x), \psi(y) \rangle &= \sum_{j=0}^\infty 2^{-(j+1)} \sum_{(k_1,\dots,k_j)\in \N^j} x_{k_1}\dots x_{k_j} y_{k_1}\dots y_{k_j}.
\end{align}
The inner term on the right-hand side of Eq.~\eqref{eqn:psi-product} can be simplified to
\begin{align}\label{eqn:simplify-sum-product-terms}
\sum_{(k_1,\dots,k_j)\in \N^j} x_{k_1}\dots x_{k_j} y_{k_1}\dots y_{k_j} = (\langle x, y \rangle)^j.
\end{align}
Combining Eqs.~\eqref{eqn:psi-product} and~\eqref{eqn:simplify-sum-product-terms} and using the fact that $\langle x, y \rangle \leq 1$, we have
\begin{align*}
\langle \psi(x), \psi(y) \rangle &= \sum_{j=0}^\infty 2^{-(j+1)} (\langle x, y \rangle)^j = \frac{1}{2 - \langle x, y \rangle} = K(x,y),
\end{align*}
which verifies that $K$ is a kernel function and $\psi$ is the associated mapping. Since $\psi$ maps from $\R^\N$ to $\R^\N$ and $\ltwos{x}\leq 1$ implies $\ltwos{\psi(x)} = K(\psi(x),\psi(x)) \leq 1$, we can recursively define a sequence of mappings
\[
	\psi^{(0)}(x) = x \quad \mbox{and} \quad \psi^{(p)}(x) = \psi(\psi^{(p-1)}(x)).
\]
Using the relation between $K$ and $\psi$, it is easy to verify that the associated kernels are
\begin{align}\label{eqn:recursive-kernel}
	K^{(0)}(x,y) = \langle x,y\rangle \quad \mbox{and} \quad K^{(p)}(x,y) = \frac{1}{2 - K^{(p-1)}(x,y)},
\end{align}
which satisfy $\langle \psi^{(p)}(x), \psi^{(p)}(y)\rangle = K^{(p)}(x,y)$.
Thus, the kernel function $K^{(k)}(x,y)$ can be easily computed from the inner product of $x$ and $y$.

\subsection{Algorithm}
\label{sec:algorithm}

\begin{algorithm}[t]
\DontPrintSemicolon
\KwIn{Feature-label pairs $\{(x_i,y_i)\}_{i=1}^n$; Loss function $\ell:\R\times \R\to \R$; Number of hidden layers $k$; Regularization coefficient $B$.}

Solve the following convex optimization problem:
\[
	\widehat\alpha = \arg\min_{\alpha\in \R^n} \frac{1}{n}\sum_{j=1}^n \ell\left(\sum_{i=1}^n \alpha_i K^{(k)}(x_i,x_j), y_i\right)\quad \mbox{s.t.}\quad 
	\sum_{i,j=1}^n \alpha_i \alpha_j K^{(k)}(x_i, x_j) \leq B^2
\]
where $K^{(k)}$ is defined in Eq.~\eqref{eqn:recursive-kernel}.

\KwOut{Predictor $\fhat_n(x) = \sum_{i=1}^n \widehat \alpha_i K^{(k)}(x_i,x)$.}
\caption{Recursive Kernel Method for Learning Neural Network}
\label{alg:kernel-method}
\end{algorithm}

We are now ready to specify the algorithm to learn the neural network. Suppose that the neural network has $k$ hidden layers. Let $\F_k$ represent the Reproducing Kernel Hilbert Space (RKHS) induced by the kernel $K^{(k)}$ and let $\F_{k,B} \subset \F_k$ be the set of RKHS elements whose norm are bounded by $B$. Given training examples $\{(x_i,y_i)\}_{i=1}^n$, define the predictor
\[
	\fhat_n \defeq \arg\min_{f\in \F_{k,B}} \frac{1}{n}\sum_{i=1}^n \ell(f(x_i),y_i).
\]
According to the representer theorem, we can represent $\fhat_n$ by
\begin{align}\label{eqn:predictor-family}
	\fhat_n(x) = \sum_{i=1}^n \alpha_i K^{(k)}(x_i, x) \quad \mbox{where}\quad  \sum_{i,j=1}^n \alpha_i \alpha_j K^{(k)}(x_i, x_j) \leq B^2,
\end{align}
Computing the vector $\alpha$ is a convex optimization problem in $\R^n$ and therefore can be solved in time $\poly(n,d)$ using standard optimization tools. We call this algorithm the \emph{recursive kernel method} and summarize it in Algorithm~\ref{alg:kernel-method}. It is an improper learning algorithm since the learned predictor $\fhat_n$ cannot be represented by a neural network.

\subsection{Main Result}

Applying classical results from learning theory, we can upper bound the
Rademacher complexity of $\F_{k,B}$ by $\sqrt{2B^2/n}$ (see, e.g.,~\cite{kakade2009complexity}). Thus, with probability at least $1-\delta$, we can upper bound the generalization loss of predictor $\fhat_n(x)$ by
\[
	\E[\ell(\fhat_n(x),y)] \leq \arg\min_{f\in \F_{k,B}} \E[\ell(f(x),y)] + \epsilon,
\]
when the sample size $n = \Omega(B^2\log(1/\delta)/\epsilon^2)$. See~\cite[][Theorem 2.2]{shalev2011learning} for the proof of this claim. In order to establish the bound~\eqref{eqn:epsilon-optimality-bound}, it suffices to show that $\nn_{k,L,\sigma} \subset \F_{k,B}$ where $B$ is a constant that only depends on $k$ and $L$. The following lemma establishes the claim. See Appendix~\ref{sec:proof-lemma-family-contains-nn} for the proof.

\begin{lemma}\label{lemma:family-contains-nn}
Assume that the function $\sigma(x)$ has a polynomial expansion $\sigma(x) = \sum_{j=0}^\infty \beta_j x^j$. Let $H(\lambda) \defeq L\cdot \sqrt{\sum_{j=0}^\infty 2^{j+1} \beta_j^2 \lambda^{2j}}$ and define $H^{(k)}(x)$ be the degree-$k$ composition of function $H$, then
$\nn_{k,L,\sigma} \subset \F_{k, H^{(k)}(L)}$.
\end{lemma}

\noindent Using Lemma~\ref{lemma:family-contains-nn} and the above analyses, we obtain the main result of this paper.

\begin{theorem}\label{theorem:upper-bound}
Let Assumption~\ref{assu:boundedness} be true and define $F(k,L) \defeq H^{(k)}(L)$ where $H^{(k)}(L)$ is specified in Lemma~\ref{lemma:family-contains-nn}. If $F(k,L)$ is finite, then with probability at least $1-\delta$, the predictor defined in Algorithm~\ref{alg:kernel-method} achieves
\[
	\E[\ell(\fhat_n(x),y)] \leq \arg\min_{\nn \in \nn_{k,L,\sigma}}\E[\ell(\nn(x),y)] + \epsilon.
\]
The sample complexity is bounded by $\poly(1/\epsilon,\log(1/\delta),F(k,L))$; the time complexity is bounded by $\poly(d,1/\epsilon,\log(1/\delta),F(k,L))$.
\end{theorem}

\begin{figure}
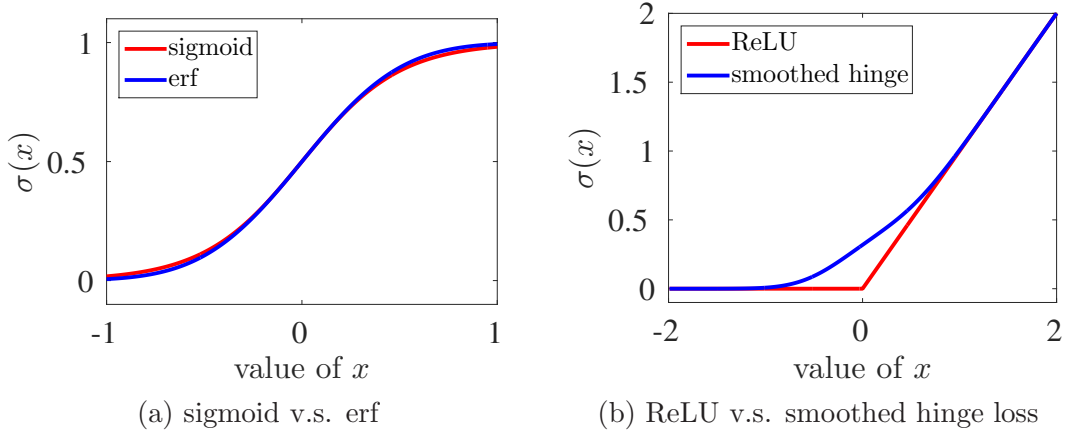

\centering
\begin{tabular}{ccc}
\includegraphics[width = 0.4\textwidth]{sigmoid-func} & &
\includegraphics[width = 0.4\textwidth]{relu-func}\\
(a) sigmoid v.s. erf && (b) ReLU v.s. smoothed hinge loss
\end{tabular}
\caption{Comparing different activation functions. The two functions in (a) are quite similar. The smooth hinge loss in (b) is a smoothed version of ReLU.}\label{fig:compare-activation}
\end{figure}

\subsection{Examples}
\label{sec:example}

We study several concrete examples where $F(k,L)$ is finite.
Our first example is the quadratic activation function:
\[
	\sq(x) = x^2.
\]
This activation function has been studied by Livni et al.~\cite{livni2014computational}, who refer to a neural network activated by this function as a polynomial network. In Theorem~\ref{theorem:upper-bound}, if the quadratic activation function is employed, we have $H(\lambda) = 2L\lambda^2$. As a consequence, we have $F(1,L) = 2L^2$ and more generally $F(k,L) \leq (2L)^{2^{k+1}-1}$ by induction. Thus, the sample and the time complexity of Algorithm~\ref{alg:kernel-method} is a polynomial function of $(d,1/\epsilon, \log(1/\delta), L)$ for any constant $k$.

Next, we study sigmoid-like or ReLU-like activation functions. We consider a shifted erf function defined as:
\[
	\erf(x) = \frac{1}{2}(1 + {\rm erf}(\sqrt{\pi}x)),
\]
and a smoothed hinge loss function defined as:
\[
	\sh(x) = \int_{-\infty}^x \erf(t)dt = \erf(x) \cdot x + \frac{e^{-\pi x^2}}{2\pi}.
\]
In Figure~\ref{fig:compare-activation}, we compare $\erf$ and $\sh$ with the sigmoid function and the ReLU function. It is seen that $\erf$ is similar to the sigmoid function and $\sh$ is a smoothed version of ReLU. It is also easy to verify that $\erf$ is sigmoid-like and $\sh$ is ReLU-like. The following proposition shows that if either $\erf$ or $\sh$ is used as the activation function, the quantity $F(k,L)$ is finite. See Appendix~\ref{sec:proof-lemma-func-expansion} for the proof.

\begin{proposition}\label{lemma:func-expansion}
For the $\erf$ function, we have 
\[
	H(\lambda) \leq L\cdot \sqrt{\frac{1}{2} + 4\lambda^2(1+3 e \pi \lambda^2 e^{4\pi \lambda^2 })}  \quad \mbox{for any $\lambda \geq 3$}.
\]
For the $\sh$ function, we have 
\[
	H(\lambda) \leq L\cdot \sqrt{\lambda^2 + 8\lambda^4(1+3 e \pi \lambda^2 e^{4\pi \lambda^2 })}  \quad \mbox{for any $\lambda \geq 3$}.
\]
\end{proposition}

\noindent Thus, Theorem~\ref{theorem:upper-bound} implies that the neural network activated by $\erf$ or $\sh$ is learnable in polynomial time given any constant $(k,L)$. 

Finally, we demonstrate how the conditions of Assumption~\ref{assu:boundedness} could be modified. Consider a sigmoid-activated network with $k$ hidden layers which satisfies the following:
\[
	\sum_{j=1}^{d^{(p)}} |w_{i,j}^{(p)}| \leq L \quad \mbox{for all } (p,i) \in \{1,\dots,k\} \times \{0,\dots, d^{(p+1)}\}.
\]
This means that the $\ell_1$-norm of all layers is bounded by $L$. In addition, we assume that the input vector satisfies $\|x\|_\infty \leq 1$. This is in contrast to the condition $\|x\|_2 \leq 1$ in Assumption~\ref{assu:boundedness}. It was shown by Livni et al.~\cite[][Theorem 4]{livni2014computational} that this sigmoid network can be approximated by a polynomial network with arbitrarily small approximation error $\epsilon$. The associated polynomial network has $\order(k\log(Lk+L\log(1/\epsilon)))$ hidden layers, whose $\ell_1$-norms are bounded by $e^{\order(L\log(1/\epsilon))}$. If we normalize the input vector $x\in \R^d$ by $x \leftarrow x/\sqrt{d}$ and multiple all first-layer weights by $\sqrt{d}$, the output of the network remains invariant and it satisfies Assumption~\ref{assu:boundedness}. Thus, combining our result for the polynomial network and the above analysis, the sigmoid network can be learned in
\[
	\poly\left(d^{(Lk+L\log(1/\epsilon))^{\order(k)}}, \log(1/\delta)\right)
\]
sample and time complexity. This is a quasi-polynomial dependence on $1/\epsilon$ for any constant $(k,L)$. Notice that the dimension $d$ comes into the expression.

\section{Hardness Result}
\label{sec:hardness}

In Section~\ref{sec:example}, we see that the dependence of the time complexity on $L$ is at least exponential for $\erf$ and $\sh$, but it is polynomial for the quadratic activation. It is thus natural to wonder if there is a sigmoid-like or ReLU-like activation function that makes the time complexity a polynomial function of $L$. In this section, we prove that this is impossible given standard hardness assumptions. 

Our proof relies on the hardness of standard (nonagnostic) PAC learning of intersection of halfspaces given in Klivans and Sherstov~\cite{klivans2006cryptographic}. More precisely, let 
\[
	H = \{ x\to {\rm sign}(w^T x - b - 1/2): x\in\{-1,1\}^d,~b\in \N, ~w\in \N^d,~|b| + \lones{w} \leq \poly(d) \}
\]
be the family of halfspace indicator functions mapping $\mathcal{X} = \{-1,1\}^d$ to $\{-1,1\}$, and let $H_T$ be the set of functions taking the form:
\[
	h(x) = \left\{ \begin{array}{cl}
	1 & \mbox{if $h_1(x) = \dots = h_T(x) = 1$},\\
	-1 & \mbox{otherwise}.
	\end{array}\right. \quad \mbox{where $h_1,\dots,h_T\in H$}.
\]
Thus, $H_T$ is the set of functions that indicates the intersection of $T$ halfspaces.
For any distribution on $\mathcal{X}$, an algorithm $\mathcal{A}$ takes a sequence of $(x,h^*(x))$ as input where $x$ is a sample from $\mathcal{X}$ and $h^*\in H_T$. The algorithm learns a function $\hhat$ such that with probability at least $1-\delta$, one has
\begin{align}\label{eqn:halfspace-error-bound}
	P(\hhat(x) \neq h^*(x)) \leq \epsilon.
\end{align}
If there is such an algorithm $\mathcal{A}$ whose sample complexity and time complexity scale as $\poly(d)$, then we say that $H_T$ is efficiently learnable. Klivans and Sherstov~\cite{klivans2006cryptographic} show that $H_T$ is not efficiently learnable under a certain cryptographic assumption.

\begin{theorem}[Klivans and Sherstov~\cite{klivans2006cryptographic}]\label{theorem:halfspace}
If $T = d^\rho$ for some constant $\rho > 0$, then under a certain cryptographic assumption, $H_T$ is not efficiently learnable.
\end{theorem}

We use this hardness result to prove the hardness of learning neural networks. In particular, we construct a neural network $\nn$ such that
if there is a learning algorithm computing a predictor $\fhat$ such that $\E[\ell(\fhat(x),y)] \leq \E[\ell(\nn(x),y)] + \epsilon$,
then the error bound~\eqref{eqn:halfspace-error-bound} is satisfied. Thus, the hardness of learning intersection of halfspaces implies the hardness of learning neural networks. See Appendix~\ref{sec:proof-theorem-lower-bound} for the proof.

\begin{theorem}\label{theorem:lower-bound}
Assume the  cryptographic assumption of Theorem~\ref{theorem:halfspace}. Let $\sigma$ be a sigmoid-like or ReLU-like function and let $\ell(f(x),y) = \max(0, 1 - yf(x))$ be the hinge loss. For fixed $(\delta,\epsilon)$, there is no algorithm running in $\poly(L)$ time that learns a predictor $\fhat$ satisfying
\begin{align}\label{eqn:lower-bound-expr}
	\E[\ell(\fhat(x),y)] \leq \arg\min_{\nn \in \nn_{1,L,\sigma}}\E[\ell(\nn(x),y)] + \epsilon \quad \mbox{with probability at least $1-\delta$}.
\end{align}
\end{theorem}

The hardness of learning sigmoid-activated and ReLU-activated neural networks has been proved by Livni et al.~\cite{livni2014computational} when $\ell$ is the zero-one loss. Theorem~\ref{theorem:lower-bound} presents a more general result, showing that any activation function that is sigmoid-like or ReLU-like leads to the computational hardness, even if the loss function $\ell$ is convex. 

\section{Experiments}
\label{sec:experiment}

In this section, we compare the proposed algorithm with several baseline algorithms on the MNIST digit recognition task. Since the basic MNIST digits are relatively easy to classify, we introduce three variations which make the problem more challenging.

\begin{figure}
\centering
\begin{tabular}{ccc}
\includegraphics[width = 0.053\textwidth]{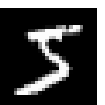}~
\includegraphics[width = 0.053\textwidth]{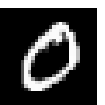}~
\includegraphics[width = 0.053\textwidth]{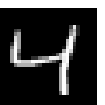}~
\includegraphics[width = 0.053\textwidth]{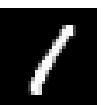}~
\includegraphics[width = 0.053\textwidth]{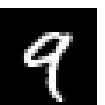}~
\includegraphics[width = 0.053\textwidth]{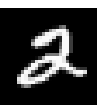}
&&
\includegraphics[width = 0.4\textwidth]{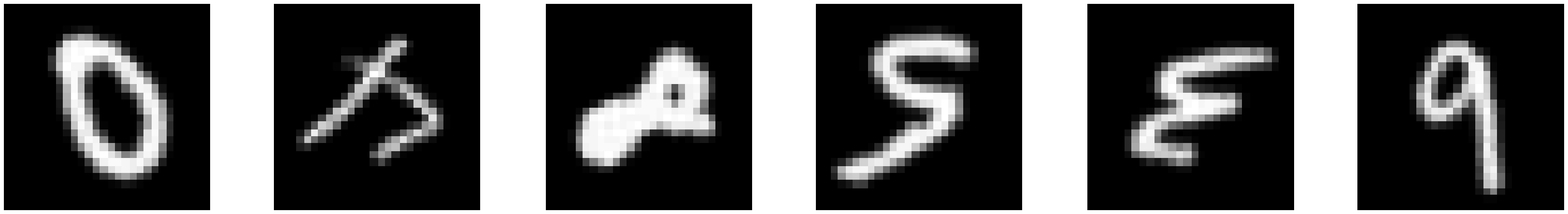}\\
(a) Basic && (b) Rotation\\
&&\\
\includegraphics[width = 0.4\textwidth]{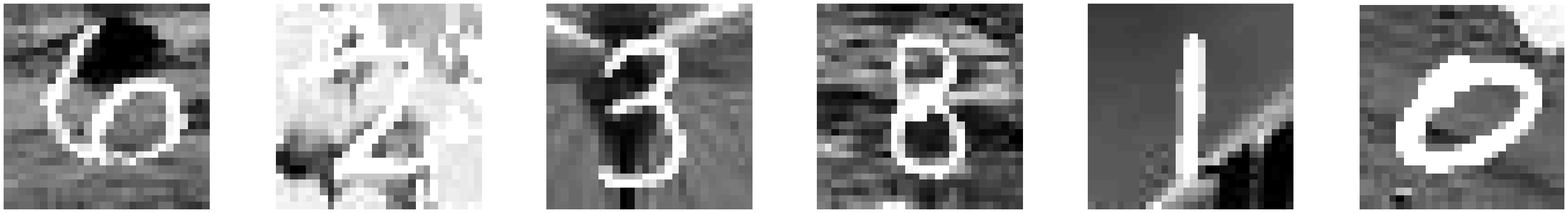}&&
\includegraphics[width = 0.4\textwidth]{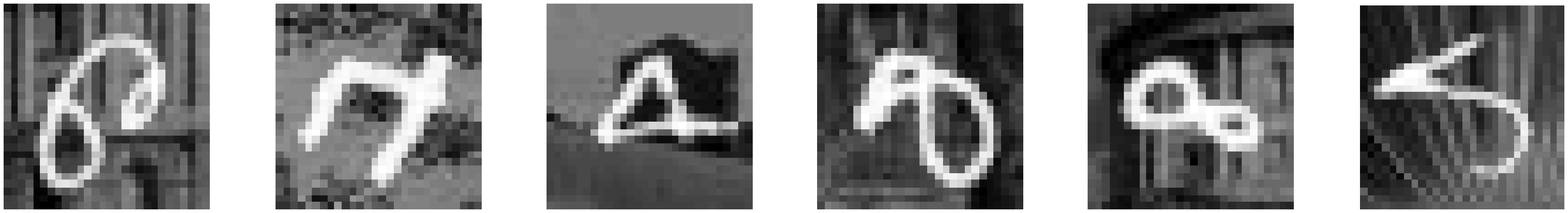}\\
(c) Background && (d) Background + Rotation
\end{tabular}
\caption{The MNIST dataset and its variations. (a) the basic MNIST dataset; (b) the digits were rotated by an angle generated uniformly between $0$ and $2\pi$. (c) a black and white image was used as the background for the digit image; (d) the background perturbation and the rotation perturbation are combined.}\label{fig:mnist}
\end{figure}

\paragraph{Datasets} We use the MNIST handwritten digits dataset and three variations of it. See Figure~\ref{fig:mnist} for the description of these datasets and several exemplary images. All the images
are of size $28\times 28$. For all datasets, we use 10,000 images for training, 2,000 images for validation and 50,000 images for testing. 
This partitioning is recommended by the source of the data~\cite{WinNT}.

\paragraph{Algorithms} For the recursive kernel method, we train one-vs-all SVM classifiers with Algorithm~\ref{alg:kernel-method}. The hyper-parameters are given by $k \in \{1,4\}$ and $B = 100$. All images are pre-processed by the following steps: deskewing, centering and normalization. The deskewing step computes the principal axis of the shape that is closest to the vertical, and shifts the lines so as to make it vertical. It is a common preprocessing step for the kernel method~\cite{lecun1998gradient}. The centering and normalization steps center the feature vector and scale it to have the unit $\ell_2$-norm. 

We compare with the following baseline models: multi-class logistic regression, multi-layer perceptron and convolution neural networks. The multi-layer perceptron is a fully connected neural network with a single hidden layer which contains 500 hidden neurons. It covers the networks that can be learned by the method of Janzamin et al.~\cite{janzamin2015generalization}. The convolution neural networks implement the LeNet5 architecture~\cite{lecun1998gradient}. All baseline models are trained via stochastic gradient descent.

\begin{table}
\centering
\begin{tabular}{|c|r|r|r|r|}
\hline
	&Basic	& Rotation	& Background & Background+Rotation\\\hline
Logistic Regression	& 9.53\% &	46.01\%	&28.05\% &	66.93\%\\
Multilayer Perceptron	&4.98\%	& 14.72\%&	28.68\% &	63.91\%\\
LeNet5 &{\bf 2.08\%}	&9.27\%	& {\bf 9.35\%}	& {\bf 32.36\%} \\\hline
Recursive Kernel ($k = 1$)	& 3.31\%	 & 9.71\%	&22.39\% &	53.72\%\\
Recursive Kernel ($k = 4$)	& 3.08\%	 & {\bf 8.78\%}	&22.13\%	 &   52.94\%\\\hline
\end{tabular}
\caption{Classification error rates of different methods on the MNIST dataset and its variations. The best results are marked by the bold font.}\label{table:mnist-error}
\end{table}

\paragraph{Results} The classification error rates  are summarized in Table~\ref{table:mnist-error}. As the table shows, the recursive kernel method is consistently more accurate than logistic regression and the multi-layer perceptron. On the Basic and the Rotation datasets, the proposed algorithm is comparable with LeNet5. On the other two datasets, LeNet5 wins over other methods by a relatively large margin. It is worth noting that when we choose a greater $k$, the performance of the proposed algorithm gets better. Recall that a greater $k$ learns a deeper neural network, thus the empirical observation is intuitive. 

Although the recursive kernel method doesn't outperform the LeNet5 model, the experiment demonstrates that it does learn better predictors than fully connected neural networks such as the multi-layer perceptron. The LeNet5 architecture encodes prior knowledge about digit recogniition via the convolution and pooling operations; thus its performance is better than the generic architectures.

\section{Conclusion}

In this paper, we have presented an algorithm and a theoretical analysis for the improper learning of multi-layer neural networks. The proposed method, which is based on a recursively defined kernel, is guaranteed to learn the neural network if it has a constant depth and a constant $\ell_1$-norm. We also present hardness results showing that the time complexity cannot be polynomial in the $\ell_1$-norm bound. We compare the algorithm with several baseline methods on the MNIST dataset and its variations. The algorithm learns better predictors than the full-connected multi-layer perceptron but is outperformed by LeNet5. We view this line of work as a contribution to the ongoing effort to develop learning algorithms for neural networks that are both understandable in theory and useful in practice.

\bibliographystyle{abbrvnat} \bibliography{bib}
\newpage
\appendix

\section*{Appendix}
\section{Proof of Lemma~\ref{lemma:family-contains-nn}}
\label{sec:proof-lemma-family-contains-nn}

Consider an arbitrary neural network $\nn\in \nn_{k,L,\sigma}$. Let $g^{(p)}_i \defeq \sum_{j=1}^{d^{(p)}} w^{(p)}_{ji} y^{(p)}_j$ represent the input of the neuron $i$ at layer $p+1$. Note that $g^{(p)}_i$ is a function of the input vector $x$. By this definition, it suffices to show that $g^{(k)}_1\in \F_{k, H^{(k)}(L)}$.

We claim that $g^{(p)}_i\in \F_{p, H^{(p)}(L)}$ for any $p\in \{0,1,\dots,k\}$ and prove the claim by induction. For $p = 0$, we have
\[
	g^{(0)}_i(x) = \sum_{j=1}^{d} w^{(0)}_{i,j} x_j = \langle w^{(0)}_i, \psi^{(0)}(x)\rangle.
\]
Thus, $g^{(0)}_i$ belongs to the RKHS induced by the kernel $K^{(0)}$. Furthermore, we have $\norms{g^{(0)}_i}_{\F_0} = \ltwos{w^{(0)}_i} \leq L = H^{(0)}(L)$, which implies $g^{(0)}_i\in \F_{0,H^{(0)}(L)}$. 

For $p > 0$, we assume that the claim holds for $p-1$ and we will prove it for $p$. The definition of $g^{(p)}_i$ implies
\[
	g^{(p)}_i(x) = \sum_{j=1}^{d^{(p)}} w^{(p)}_{ji} \sigma\Big( g^{(p-1)}_j(x)  \Big).
\]
Using the inductive hypothesis, we have $g^{(p-1)}_j\in \F_{p-1,H^{(p-1)}(L)}$, which implies that $g^{(p-1)}_j(x) = \langle v_j, \psi^{(p-1)}(x)\rangle$ for some $v_j\in \R^\N$, and $\ltwos{v_j} \leq H^{(p-1)}(L)$.
This implies
\begin{align}\label{eqn:gip-expansion}
g^{(p)}_i(x) &= \sum_{j=1}^{d^{(p)}} w^{(p)}_{i,j} \sigma (\langle v_j, \psi^{(p-1)}(x)\rangle).
\end{align}
Let $x^{(p-1)}$ be a shorthand notation of $\psi^{(p-1)}(x)$.
We define vector $u_j\in \R^\N$ as follow: the $(k_1,\dots,k_t)$-th coordinate of $u_j$, where $t\in \N$ and $k_1,\dots,k_t\in \N^+$, is equal to $2^{\frac{t+1}{2}} \beta_t v_{j,k_1}\dots v_{j,k_t}$. By this definition, we have
\begin{align}\label{eqn:linearize-sigma}
 \sigma(\langle v_j, x^{(p-1)} \rangle)&=  \sum_{t=0}^\infty \beta_t (\langle v_j, x^{(p-1)} \rangle)^t \nonumber\\
 &=\sum_{t=0}^\infty \beta_t \sum_{(k_1,\dots,k_t)\in \N^t} v_{j,k_1}\dots v_{j,k_t} x^{(p-1)}_{k_1}\dots x^{(p-1)}_{k_t}\nonumber\\
 &= 	\langle u_j, \psi(x^{(p-1)}) \rangle,
\end{align}
where the first equation holds since $\sigma(x)$ has a polynomial expansion $\sigma(x) = \sum_{t=0}^\infty \beta_t x^t$, the second by expanding the inner product, and the third by definition of $\psi(x)$ . Combining Eq.~\eqref{eqn:gip-expansion} and Eq.~\eqref{eqn:linearize-sigma}, we have
\begin{align*}
g^{(p)}_i(x) &= \sum_{j=1}^{d^{(p)}} w^{(p)}_{i,j} \langle u_j, \psi(\psi^{(p-1)}(x)) \rangle = \Big\langle  \sum_{j=1}^{d^{(p)}} w^{(p)}_{ji} u_j ,\psi^{(p)}(x) \Big\rangle.
\end{align*}
This implies that $g^{(p)}_i$ belongs to the RKHS induced by the kernel $K^{(p)}$. 

Finally, we upper bound the norm of $g^{(p)}_i$. Notice that
\begin{align}\label{eqn:gip-norm-bound}
\norms{g^{(p)}_i}_{\F_p} & = \Big\| \sum_{j=1}^{d^{(p)}} w^{(p)}_{i,j} u_j \Big\|_2 \leq \sum_{i=1}^{d^{(p)}} |w^{(p)}_{i,j}|\cdot \ltwos{u_j}
\leq L\cdot\max_{j\in [d^{(p)}]}\{ \ltwos{u_j} \}.
\end{align}
Using the definition of $u_j$ and the inductive hypothesis, we have
\begin{align}
\ltwos{u_j}^2 &= \sum_{t=0}^\infty 2^{t+1} \beta_t^2 \sum_{(k_1,\dots,k_t)\in \N^t}  v_{j,k_1}^2 v_{j,k_2}^2 \cdots v_{j,k_t}^2\nonumber\\
&= \sum_{t=0}^\infty 2^{t+1} \beta_t^2 \ltwos{v_j}^{2t}
\leq \sum_{t=0}^\infty 2^{t+1} \beta_t^2 (H^{(p-1)}(L))^{2t}.\label{eqn:gip-norm-induction}
\end{align}
Combining inequality~\eqref{eqn:gip-norm-bound} and~\eqref{eqn:gip-norm-induction}, we have $\norms{g^{(p)}_i}_{\F_p} \leq H^{(p)}(L)$, which verifies that $g^{(p)}_i\in \F_{p, H^{(p)}(L)}$.


\section{Proof of Proposition~\ref{lemma:func-expansion}}
\label{sec:proof-lemma-func-expansion}

For the $\erf$ function, the polynomial expansion is
\[
	\erf(x) = \frac{1}{2} + \frac{1}{\sqrt{\pi}}\sum_{j=0}^{\infty} \frac{(-1)^j(\sqrt{\pi}x)^{2j+1}}{j!(2j+1)}.
\]
Therefore, we have
\begin{align}\label{eqn:erf-h-expansion}
	H(\lambda) = L\cdot \sqrt{ \frac{1}{2} + \frac{2}{\pi}\sum_{j=0}^{\infty} \frac{(2\pi \lambda^2)^{2j+1}}{(j!)^2(2j+1)^2} }.
\end{align}
Shalev-Shwartz et al.~\cite[][Corollary C]{shalev2011learning} provide an upper bound on the right-hand side of Eq.~\eqref{eqn:erf-h-expansion}.
In particular, they prove that
\begin{align}\label{eqn:ss-bound}
	\frac{2}{\pi}\sum_{j=0}^{\infty} \frac{(2\pi \lambda^2)^{2j+1}}{(j!)^2(2j+1)^2} \leq 4\lambda^2(1+3 e \pi \lambda^2 e^{4\pi \lambda^2 }) \quad \mbox{for any $\lambda \geq 3$}.
\end{align}
Plugging this upper bound to Eq.~\eqref{eqn:erf-h-expansion} completes the proof.\\

For the $\sh$ function, since it is the integral of the $\erf$ function, its polynomial expansion is
\[
	\sh(x) = \frac{x}{2} + \frac{1}{\sqrt{\pi}}\sum_{j=0}^{\infty} \frac{(-1)^j(\sqrt{\pi}x)^{2j+1}x}{j!(2j+1)(2j+2)},
\]
and consequently,
\begin{align}\label{eqn:sh-h-expansion}
	H(\lambda) = L\cdot \sqrt{ \lambda^2 + \frac{2}{\pi}\sum_{j=0}^{\infty} \frac{(2\pi \lambda^2)^{2j+1}(2\lambda^2)}{(j!)^2(2j+1)^2(2j+2)^2 } }. 
\end{align}
We upper bound the right-hand side of Eq.~\eqref{eqn:sh-h-expansion} by
\begin{align*}
\frac{2}{\pi}\sum_{j=0}^{\infty} \frac{(2\pi \lambda^2)^{2j+1}(2\lambda^2)}{(j!)^2(2j+1)^2(2j+2)^2} &\leq \frac{4\lambda^2}{\pi}\sum_{j=0}^{\infty} \frac{(2\pi \lambda^2)^{2j+1}}{(j!)^2(2j+1)^2} \\
&\leq 8\lambda^4(1+3 e \pi \lambda^2 e^{4\pi \lambda^2 }) \quad \mbox{for any $\lambda \geq 3$},
\end{align*}
where the final inequality holds because of Eq.~\eqref{eqn:ss-bound}.
Plugging this upper bound into Eq.~\eqref{eqn:sh-h-expansion} completes the proof.


\section{Proof of Theorem~\ref{theorem:lower-bound}}
\label{sec:proof-theorem-lower-bound}

We construct a one-hidden-layer neural network that encodes the intersection of $T$ halfspaces. Suppose that the $t$-th halfspace is characterized by $g_t(x) = w_t^T x - b_t - 1/2$. Since both $x$, $w_t$ and $b_t$ are composed of integers, we have $g_t(x) \geq 1/2$ when $h_t(x) = 1$, and $g_t(x) \leq -1/2$ when $h_t(x) = -1$.  We extend $x$ to be $(x,1)$, then extend $w_t$ to be $(w_t,b_t)$, and define
\[
	\gtilde_t(x) = \langle \wtilde_t, \xtilde \rangle \quad \mbox{where} \quad
	\xtilde \defeq \frac{1}{\sqrt{d+1}}(x,1) \mbox{ and }
	\wtilde \defeq 2\lambda\sqrt{d+1}(w_t,b_t),
\]
where $\lambda$ is a scalar to be specified. According to this definition, we have $\ltwos{\xtilde} = 1$ and $\ltwos{\wtilde} = \poly(d)$. In addition, we have $\gtilde_t(x) \geq \lambda$ when $h_t(x) = 1$, and $\gtilde_t(x) \leq -\lambda$ when $h_t(x) = -1$.

\paragraph{Sigmoid-like Activation} If $\sigma$ a is sigmoid-like function, there is a constant $c$ such that 
\[
\lim_{x\to-\infty} x^c \sigma(x) = \lim_{x\to \infty} x^c (1-\sigma(x)) = 0.
\]
Thus, there is a sufficiently large constant $C$ such that $\sigma(x) \leq x^{-c}$ 
for all $x \leq -C$ and $\sigma(x) \geq 1 - x^{-c}$ for all $x \geq C$. Note that the number $T$  of intersecting halfspaces is a polynomial function of dimension $d$. As a consequence, there is a sufficiently large constant $\lambda \sim \poly(d)$ such that
\[
	\sigma(x) \geq 1 - \frac{1}{4 T} \mbox{ for all $x > \lambda$} \quad \mbox{and}\quad \sigma(x) \leq \frac{1}{4 T} \mbox{ for all $x \leq -\lambda$}.
\]
Thus, we have $\sigma(\gtilde_t(x)) \geq 1- \frac{1}{4T}$ if $h_t(x) = 1$ and $\sigma(\gtilde_t(x)) \leq \frac{1}{4T}$ if $h_t(x) = -1$.

We define the neural network $\nn$ to be
\begin{align}\label{eqn:hardness-def-nn}
	\nn(x) = \sum_{t=1}^T 4\;\sigma(\gtilde_t(x)) -(4T-2).
\end{align}
It is easy to verify that $\nn\in \nn_{1,L,\sigma}$ for some $L\sim\poly(d)$.
If $h^*(x) = 1$, then $x$ belongs to the intersection of halfspaces. It implies that $\sigma(\gtilde_t(x)) \geq 1 - \frac{1}{4 T}$ for all $t\in [T]$. Combining with Eq.~\eqref{eqn:hardness-def-nn}, we obtain $\nn(x) \geq 1$. On the other  hand, if $h^*(x) = -1$, then there is some $t$ such that $\sigma(\gtilde_t(x)) \leq \frac{1}{4 T}$. Thus, Eq.~\eqref{eqn:hardness-def-nn} implies $\nn(x) \leq -1$. In summary, we have $h^*(x)\nn(x) \geq 1$ for any $x\in \mathcal{X}$.
As a consequence, we have $\ell(\nn(x), h^*(x)) \equiv 0$ where $\ell$ is the hinge loss.

Assume that there is a predictor $\fhat$ satisfying the error bound~\eqref{eqn:lower-bound-expr}. Let
$\hhat(x) = {\rm sign}(\fhat(x))$ be a classifier that judges the intersection of hyperplanes. Since the hinge loss is an upper bound on the zero-one loss, we have
\begin{align*}
	P(\hhat(x) \neq h^*(x)) &= \E[\indicator(\hhat(x) \neq h^*(x))]
	= \E[\indicator({\rm sign}(\fhat(x))\neq h^*(x))] \leq \E[\ell(\fhat(x), h^*(x))] \\
	& \leq \E[\ell(\nn(x), h^*(x))] + \epsilon = \epsilon,
\end{align*}
where the final inequality follows from inequality~\eqref{eqn:lower-bound-expr}. The last equation holds since
$\ell(\nn(x), h^*(x)) \equiv 0$. This implies that the associated classifier $\hhat$ satisfies the error bound~\eqref{eqn:halfspace-error-bound}. Since $\hhat$ cannot be computed in $\poly(d)$ time, we conclude that $\fhat$ cannot be computed in $\poly(L)$ time.

\paragraph{ReLU-like Activation} If $\sigma$ is a ReLU-like function, then by definition, we have $\sigma'(x) \defeq \sigma(x) - \sigma(x-1)$ is a sigmoid-like function. Following the argument for the sigmoid-like activation, if we treat $\sigma'$ as the activation function, then the remaining part of the proof will go through without any further modification. This completes the proof for the ReLU-like activation.


\end{document}